\documentclass{article}
\usepackage{spconf,graphicx}
\usepackage[linesnumbered,ruled]{algorithm2e} 
\usepackage{mathtools}
\usepackage{amsmath}
\usepackage{amssymb}

\title{Image Decomposition and Classification through a Generative Model}
\name{
Houpu Yao$^{1}$, Malcolm Regan$^{2}$, Yezhou Yang$^{3}$, Yi Ren$^{1}$
}
\address{$^{1}$ Department of Mechanical and Aerospace Engineering, Arizona State University\\
    $^{2}$ Department of Electrical and Computer Engineering, North Carolina State University\\
    $^{3}$ Department of Computer Science and Engineering, Arizona State University\\
}

\begin{document}
%
\maketitle
\begin{abstract}
We demonstrate in this paper that a generative model can be designed to perform classification tasks under challenging settings, including adversarial attacks and input distribution shifts. Specifically, we propose a conditional variational autoencoder that learns both the decomposition of inputs and the distributions of the resulting components. During test, we jointly optimize the latent variables of the generator and the relaxed component labels to find the best match between the given input and the output of the generator.
The model demonstrates promising performance at recognizing overlapping components from the multiMNIST dataset, and novel component combinations from a traffic sign dataset.
Experiments also show that the proposed model achieves high robustness on MNIST and NORB datasets, in particular for high-strength gradient attacks and non-gradient attacks.
\end{abstract}
\begin{keywords}
Generative model, classification, adversarial defense 
\end{keywords}
%

\section{Introduction}
\label{sec:intro}
Neural network architectures have been developed to achieve human-level performance on standard vision tasks~\cite{ResNet,Szegedy2016Inception}. However, it is acknowledged that feedforward networks have difficulty at generalization under input distribution shifts, e.g., novel object sets~\cite{stringer2002novel} and objects with overlaps~\cite{sabour2017dynamic}. Furthermore, studies have shown that networks, even with high standard test accuracy, can suffer from imperceptible adversarial attacks~\cite{Goodfellow2014}. 
While neither distribution shifting or adversarial attacks are common cases in standard test environments for image classifiers~\cite{papernot2016practical,lu2017standard}, the demonstrated risk of existing models have raised concerns over their real-world applications (e.g., autonomous driving and security surveillance) where failed classification for a short amount of time can be catastrophic. These concerns have led to the question of whether semantic attributes or physical components of the inputs are truly understood by feedforward, albeit deep, networks~\cite{ResNet,Szegedy2016Inception}. Indeed, evidence showed that state-of-the-art classification models have a drastically different accuracy changing pattern than human beings in classifying image sequences with diminishing details~\cite{Ullman2016pnas}, suggesting that the two have different feature learning behaviors.

This concern over model generalizability and robustness is intrinsic to classifiers that perform bottom-up signal processing. Alternative models that integrate bottom-up processing with top-down reasoning through recursive inference have been studied~\cite{chen2008rapid,RCN2017}. Of particular interest are recursive compositional models (e.g., AND-OR templates~\cite{RCN2017}) that learn to match deformable objects and infer graph states for detection and recognition. These models take advantage of highly structured generators, e.g., by explicitly modeling object edges and surfaces. However, the intrinsic trade-off between model and computational complexity (for both model learning and recognition/classification) may hamper their application to general inputs for which the recognition or classification tasks depend on a richer set of features. With the rapid advance in generative models (e.g. GAN~\cite{brock2018biggan}, PixelCNN~\cite{Salimans2017PixelCNN++}, and VAE~\cite{Kingma2013,Yan}), it is tempting to investigate top-down classification mechanisms that incorporate more flexible generators than compositional models.

\begin{figure}[]
    \centering
    \includegraphics[width=\linewidth]{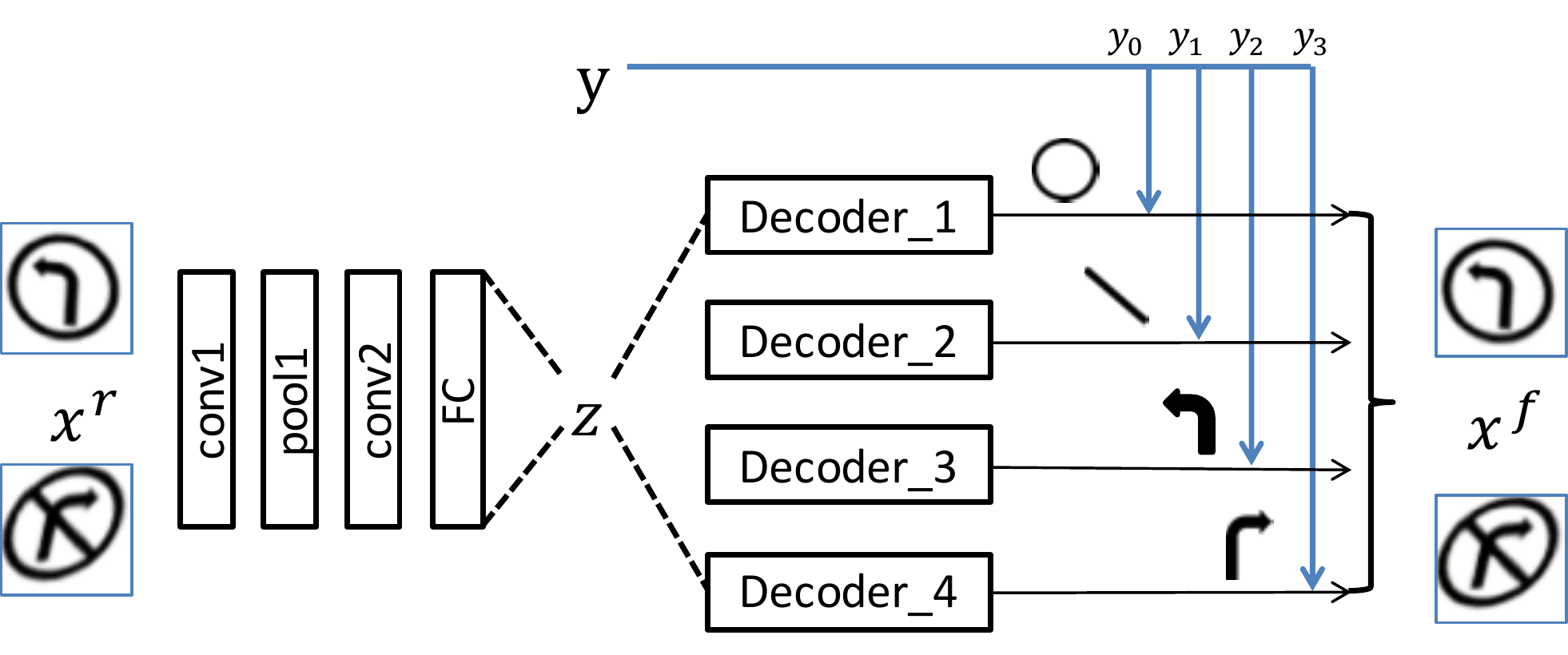}
    \caption{Schematic of the proposed generative model with sub-networks for component-wise generation. Each sub-network is gated by a binary component label $y_i$. Classification is done through optimizing $z$ and $y$ to match the generated image with the input. 
    }
    \label{fig:summary}
\end{figure}

To this end, we present in this paper a classification algorithm where input images are classified by minimizing the difference between the target image and the output of a generator. We choose Conditional Variational Autoencoder (c-VAE) as the backbone of our model due to its simplicity and good convergence property. Following the argument that attribute recognition is key to object classification~\cite{farhadi2009describing,lampert2009attribute}, we focus on the classification of attributes, or more specifically, the existence of object components from the input image, while assuming that the follow-up mapping from these components to the object class is established. As illustrated in Fig.~\ref{fig:summary}, the model is built on a variational autoencoder whose decoder is composed of separate sub-networks, where each component is associated with a sub-network. 
Experiments show that the proposed model demonstrates promising performance at recognizing overlapping components from the multiMNIST dataset, and novel component combinations from a traffic sign dataset. The model also achieves high robustness on MNIST and NORB datasets, in particular for high-strength gradient attacks and non-gradient attacks.

\section{related work}

Conditional generative models have been heavily investigated in the literature. For example, VAE structure ~\cite{Kingma2013} is extended into c-VAE ~\cite{Yan} by including attribute variables as extra inputs. Besides, the latent variables are disentangled into two sub-networks in ~\cite{Yan} to perform foreground and background separation. In our work, we further extended the c-VAE idea to have more sub-networks where each sub-network corresponds to a component. And instead of generation or segmentation, we investigate the possibility to use the conditional generative model to perform robust classification. 

The utility of generative model as a defense mechanism has been explored very recently.
For example, defense GAN ~\cite{samangouei2018defenseGAN} finds the closest generation from a GAN to a given input image, and feeds the generated image to a vanilla classifier for prediction.
Similar to our approach, Schott et al.~\cite{ICLR_VAE} chose VAE to perform image generation and used the reconstruction error as a metric to perform classification directly, without feedforward classifier involved.
Compared with ~\cite{ICLR_VAE}, 
the proposed model has an optimized network structure and classification workflow, which makes it able to learn object decomposition and handle more tasks like novel or overlapping object recognition.


\section{Proposed Method}
\label{sec:proposed}

\subsection{Customized conditional variational autoencoder}
We customize a conditional variational autoencoder to learn the decomposition of object components. Let $x^r$ and $y$ be the input images and their corresponding components, respectively. A binary component vector $y$ encodes the existence of components within $x^r$. It is assumed that the maximum amount of possible components in the dataset is $n$. Let $z = Enc (x^r,\theta) \in \mathbb{R}^p$ be the latent variable generated from the encoder $Enc(\cdot,\theta)$ with parameters $\theta$, and $x^f(z,y,\phi) = \sum_i^n y_i Dec_i(z_i, \phi_i)$ be the generated images from the set of decoders $Dec_i(\cdot, \phi_i)$ with parameters $\phi_i$. $z_i  \in \mathbb{R}^{p/n}$ is a segment of $z$ corresponding to the $i$th component. Following the formulation of variational autoencoders, we define our training loss as:
\begin{equation}
    \mathbb{E}_{\{x^r,y\} \sim \mathcal{D}} \left[||x^f(z,y,\phi)-x^r||^2_2     + C  \sum_i^n y_i^T KL(z_i)\right]
\end{equation}
where $KL(z_i)$ is the sub-network-wise KL divergence of the distribution of $z_i$ from a standard normal distribution.
In the presented experiments, Adam optimizer with a learning rate of 0.001 is used for training.

\subsection{Classification}
Based on the learned generative model, the classification process can be cast into an inverse problem, where we jointly find $z^*$ and $y^*$ that minimize the difference between the generated image $x^f$ and the target image $x^r$ in the image space:
\begin{equation}
\label{eq_opt_vallina}
\begin{split}
    &z^*,y^* = \text{argmin}_{z \in \mathbb{R}^p,y\in \{0,1\}^n} f(z,y, x^r),\\
    &f(z,y, x^r)  \equiv \lVert x^f(z,y,\phi) - x^r \rVert_2^2.
\end{split}
\end{equation}
Note that Eq.~\eqref{eq_opt_vallina} is a combinatorial problem, since $y$ is a binary vector. To make this problem feasible, we relax component labels to be continuous within $[0,1]$, so that the optimization problem becomes differentiable and can be solved efficiently through gradient descent and back-propagation. 
In addition, a $L_1$ penalty is introduced to regularize non-zero components with weight $c$. The regularization is needed so that a component will be correctly discarded (i.e., sub-network outputs suppressed) if its addition to the generated image does not improve the reconstruction quality significantly.  
Lastly, it is also found that minimizing the image-wise difference after a sigmoid transformation over the reconstruction further improves the defense performance.
With these modifications, the classifier solves the following problem:
\begin{equation}
    \begin{split}
        &z^*,y^* = \text{argmin}_{z \in \mathbb{R}^p,y\in [0,1]^n} F(z,y,x^r)\\ 
        &F(z,y,x^r) = \lVert Sig(x^f(z,y,\phi)) - Sig(x^r) \rVert_2^2  + c \lVert y \rVert_1,
    \end{split}
    \label{eq_opt}
\end{equation}
where $Sig(x) = \text{sigmoid} \big( \beta \cdot x + b \big)$, and $\beta$, $b$ are hyper-parameters.

Upon convergence, $z^*$ and $y^*$ obtained via Eq.~\eqref{eq_opt} will be processed to derive the final classification result using the following logic flow: First, if the lowest reconstruction loss found is larger than a pre-set threshold $l$, the input image will be classified as noise. This step is able to filter out adversarial attacks using non-gradient methods such as Neuroevolution of Augmenting Topologies (NEAT)~\cite{nguyen2015deep}. After this initial filtering, we apply two thresholds $y_l$ and $y_u$ ($y_{l}<y_{u}$) to $y^*$: an element $y^*_i$ is set to $0$ if $y^*_i < y_{l}$, and 1 if $y^*_i > y_{u}$. For the remaining elements of $y^*$ between $y_{l}$ and $y_{u}$, we will enumerate over all binary combinations of this subset to generate a candidate set $\mathcal{Y}$ for fine-tuning the prediction. Specifically, we compute $z^{\dagger}(y) = \text{argmin}_{z \in \mathbb{R}^p} f(z,y,x^r)$ for $y \in \mathcal{Y}$ and set the classification result as $y^{\dagger} = \text{argmin}_{y \in \mathcal{Y}} f(z^{\dagger}(y),y,x^r)$. All hyper-parameters ($\beta$,$b$,$l$,$y_l$, and $y_u$) are determined based on a validation set, and are therefore dataset dependent. 

\subsection{Implementation details}
\label{sec:implementation}
The proposed model will be tested on MNIST, NORB~\cite{lecun2004norb}, multiMNIST~\cite{sabour2017dynamic}, and a Traffic Sign (TS) dataset. For \textbf{NORB}, similar to ~\cite{sabour2017dynamic}, we downscale the image resolution to 48-by-48. The \textbf{multiMNIST} dataset is synthesized by stacking two MNIST images into an image of resolution 36-by-36, which will result in 80\% overlap between two digits on average. Every image label in multiMNIST has 2 ones  and 8 zeros. A visualization of the multiMNIST dataset can be found in  Fig.~\ref{fig:multiMNIST}. 
The \textbf{TS dataset} is prepared with affine transformation to mimic traffic signs at different view angles. There are four different types of images in the dataset: ``left turn", ``right turn", ``no left turn", and ``no right turn". These images are in a resolution of 64-by-64, with examples visualized in Fig.~\ref{fig:novel}. There are four components for TS images to represent the components: circle, slash, left arrow, and right arrow. For example, a ``No right turn" image has label $[1,1,0,1]$ since it has circle, slash, and right arrow. 

\begin{figure}[h]
    \centering
    \includegraphics[width=\linewidth]{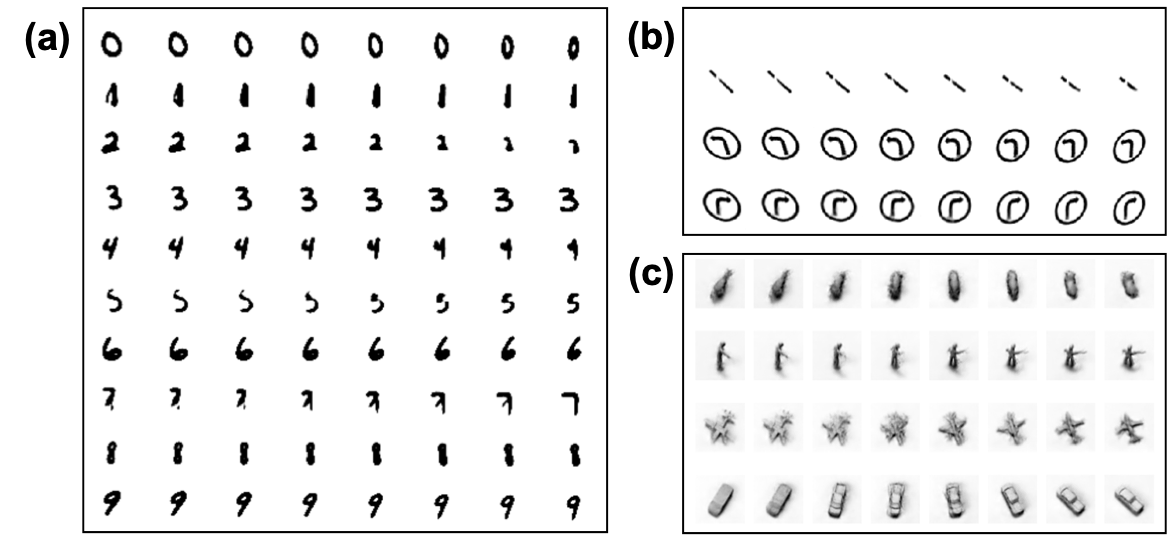}
    \caption{Sampling the latent space of each generator trained on (a) multiMNIST, (b) TS, and (c) NORB. Each sub-network is dedicated to one component. Samples of multiMNIST training images can be found in Fig.\ref{fig:multiMNIST}.}
    \label{fig:decomposition}
\end{figure}

The number of sub-networks is equal to the number of total components in the dataset, which is 10 for MNIST and multiMNIST, 5 for NORB, and 4 for TS. 
Once the network has been trained, it is expected to decompose the input object into different components, as illustrated in Fig.\ref{fig:summary}. To demonstrate the efficacy of the learning of decomposition, we visualize the latent space of each sub-networks trained on different datasets in Fig.\ref{fig:decomposition}. One observation from the results is that the model is able to remove data redundancy automatically, which can be seen from the TS dataset: To recall, the first component of a TS image represents the existence of a circle. Since the circle exists in all TS images, the learned generative model combines the circle into the generation of the arrows and leaves the first sub-network blank.

\section{Results and Discussion}
\label{sec:Results}
In this section, we demonstrate the efficacy of our method through a set of challenging experiments. These include classifying overlapping and novel objects, and defense against both gradient-based (FGM~\cite{Goodfellow2014}) and non-gradient based (NEAT~\cite{nguyen2015deep}) adversarial attacks.

\subsection{Classification of novel and overlapping objects}
We first train our model on the TS dataset by only using ``left turn", ``right turn" and ``no right turn" images for training. After the model is trained, its performance is evaluated on ``No left turn" images. This experiment is set up in a way that the network is asked to recognize novel objects not included in the training set by making use of the learned decomposition of image elements.

\begin{figure}[h]
    \centering
    \includegraphics[width=0.99\linewidth]{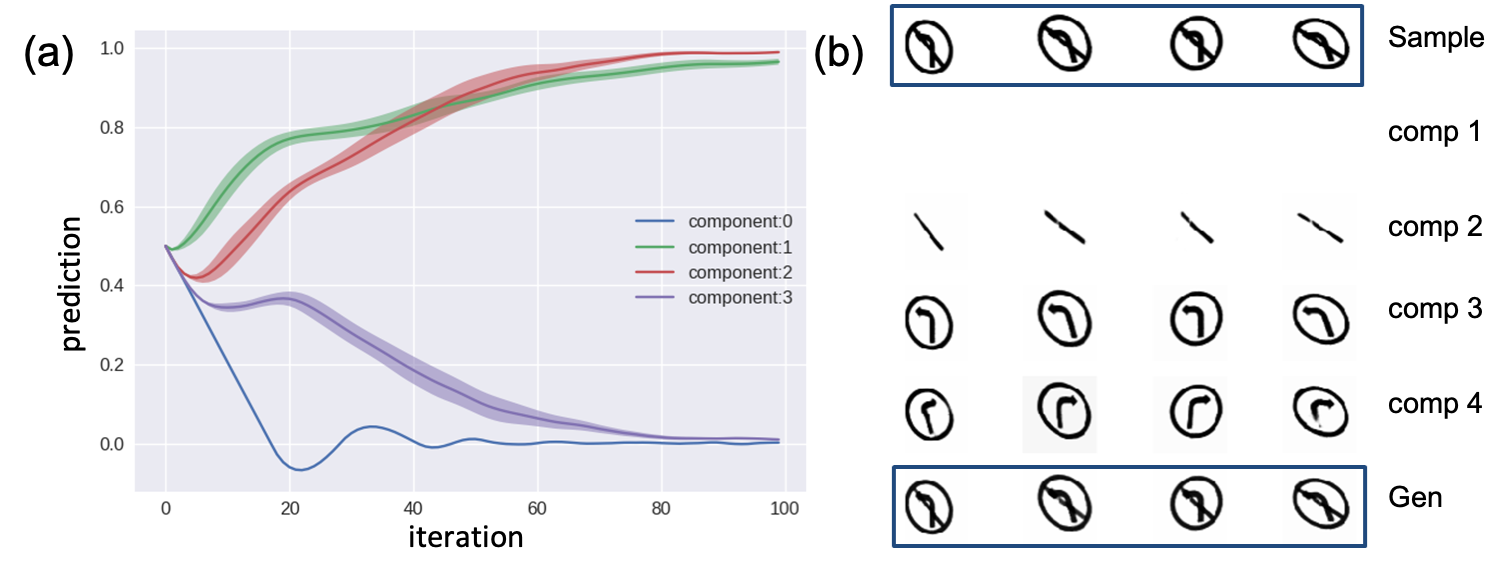}
    \caption{Classification result on novel objects. Left: convergence of component labels during the classification process, with $\pm3$ standard deviation over all ``no left turn'' test images. Right: the inputs (top), their sub-network generations (middle), and the optimal final generations (bottom).} 
    \label{fig:novel}
\end{figure}

A traditional feedforward CNN classifier will completely fail under this setting, where 99\% of the ``no left turn" images are classified as ``left turn". This is due to the fact that the classifier will associate the ``left arrow" feature with the ``left turn" category during the training phase, and this correlation strongly affects the prediction during the testing phase. In contrast, our proposed model learns to decompose the TS dataset into components after training (as shown in Fig.~\ref{fig:decomposition}b), and by combining these components together, it correctly recognizes 95\% of all novel testing images. 
The classification result is shown in Fig.~\ref{fig:novel}.  

\begin{figure}[h]
    \centering
    \includegraphics[width=0.99\linewidth]{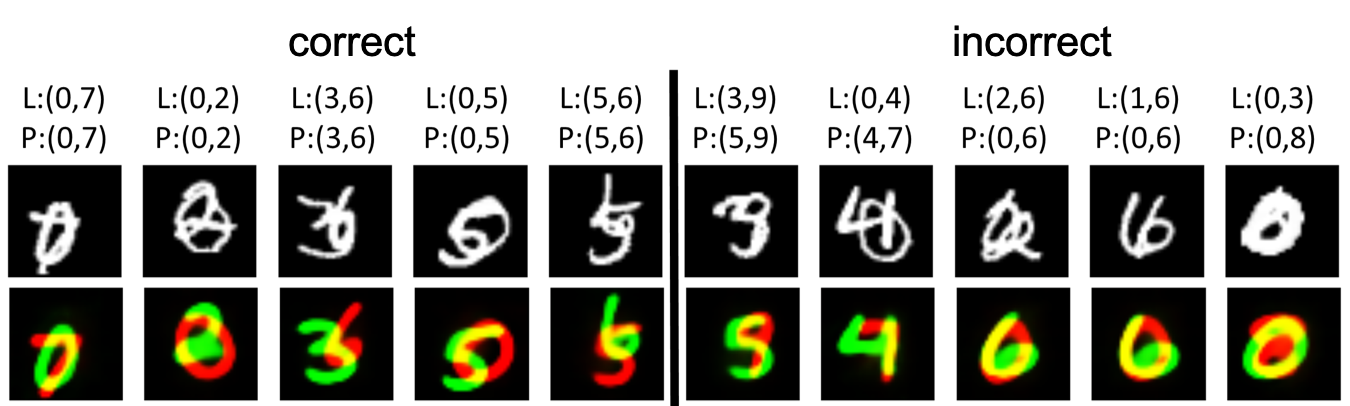}
    \caption{Examples of classification results on multiMNIST. First row: input test images. Second row: predicted images. L and P denotes ground truth and predicted labels, respectively.}
    \label{fig:multiMNIST}
\end{figure}

We now demonstrate that the proposed model is able to handle objects with large overlap using the multiMNIST dataset. 
After training, the proposed model learns the decomposition of individual digits successfully as shown in Fig.\ref{fig:decomposition}a. Applying the learned model to classifying the test dataset leads to a classification accuracy of 65.6\%. Successful and failed test samples and their classification results are shown in Fig.\ref{fig:multiMNIST}. 
While the accuracy on multiMNIST is lower than that of a CapsNet~\cite{sabour2017dynamic}(95\%), investigation of the model performance shows that many of the misclassified samples are truly difficult to be separated even for human beings. It should also be noted that the accuracy of CapsNet is resulted from 60 million training data, while our model successfully decomposes the learns the component-wise latent spaces with only 128k data points. We expect improved classification accuracy by increasing the capacity of the generative model. 


\subsection{Defending adversarial attacks}
We test the performance of our model under both gradient- and non-gradient-based adversarial attacks in this experiment. FGM and NEAT are chosen as the attack methods. 
NEAT performs evolution on Compositional Pattern Producing Networks (CPPN)~\cite{Stanley2002}, each of which represents an image. The fitness of the evolution is defined as $1-||\hat{y}-y_{target}||$, where $\hat{y}$ is the classification result of a CPPN image by the source classification model and $y_{target}$ is the target class for the attack. 
If an adversarial image is misclassified with over $90\%$ confidence, the attack is considered successful and this adversarial image will be used for test.

\begin{figure}[ht]
    \centering
    \includegraphics[width=0.99\linewidth]{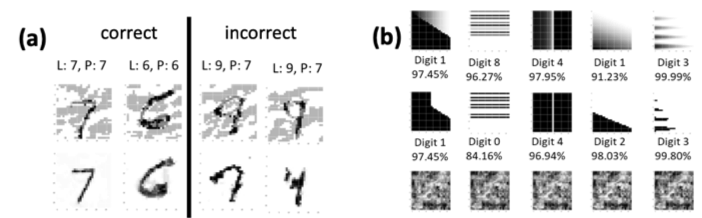}
    \caption{Sample adversarial images. Left (first and second row): FGM adversarial images and the generated images.L and P denotes ground truth and predicted labels, respectively.
    Right (from top to bottom row): CPPN adversarial images, images after binarization, and generation found by proposed method. Classification confidence of the feed-forward classifier are shown under the images.}
\label{fig:failure_cases}
\end{figure}

We first compare our method with the binarization defense~\cite{Xu2017featuresqueezing} under FGM attack in Tab.\ref{tab:FGM}. The performance of binarization drops quickly with increasing attack magnitudes, while the proposed model is able to maintain high accuracy even under relatively high attack magnitude. To further examine the cases where our model fails, we visualized adversarial MNIST samples with $\epsilon=0.4$ in Fig.\ref{fig:failure_cases}a, along with correctly classified samples. We note that many of the misclassified images are hard to be recognized even for human beings.

\begin{table}[ht]
\caption{Comparisons between baseline (binarization) and the proposed method on robustness under FGM attacks with increasing perturbation levels.}
\centering
\label{tab:FGM}
\begin{tabular}{c|c|ccccc}
\hline
 & $\epsilon$ & 0.0 & 0.1 &  0.2 & 0.3 & 0.4 \\
\hline
MNIST & baseline & \textbf{0.97} & \textbf{0.95} & 0.90 & 0.76 & 0.51\\ 
MNIST & proposed & 0.95 & 0.87 & \textbf{0.91} & \textbf{0.87} & \textbf{0.82}\\ 
\hline
NORB & baseline & 59.8 & 18.3 & 2.8 & 0.8 & 0.2 \\ 
NORB & proposed & \textbf{87.1} & \textbf{61.9} & \textbf{40.1} & \textbf{24.6} & \textbf{18.8} \\ 
\hline
\end{tabular}
\end{table}

The comparison on NEAT attacks is shown in Fig.\ref{fig:failure_cases}b.
The baseline classifier is fooled completely by the adversarial images with consistently high confidence for the targeted labels (shown in the first row of Fig.\ref{fig:failure_cases}b).
Even after binarization, many NEAT images can still be misclassified with high confidence (shown in the second row of Fig.\ref{fig:failure_cases}b).
In contrast, the proposed method is able to identify 90.6\% of these attacks. This is because the generative model is not trained on these highly structured CPPN patterns, and is not able to reconstruct them during the test. Due to the high reconstruction error, the proposed model will conclude that these input images are out of the data distribution.

\vspace{-0.3cm}
\subsection{Discussion}
While the presented experiments show promise of the proposed classification method, a deeper investigation is needed to better characterize the applicability and limitations of the proposed method to more complicated datasets (e.g., CIFAR10 and ImageNet) under more sophisticated settings (e.g. PGD adversarial attack~\cite{madry2017robust_optimization_PGD}).
Further improvements can be made to compress the generative model in order to accelerate the optimization during classification. Generative models with the capability to create fine-grained details should be studied and incorporated to further improve the performance.

\section{Conclusions}
\label{sec:conclusion}
In this paper, we investigate the utility of a tailored conditional variational autoencoder as a classifier, and test its generalizability and robustness under challenging tasks. Results show that the proposed training and classification formulations lead to promising performance: First, the model can recognize overlapping objects and novel component combinations that do not exist in the training phase. Second, our model is able to defend against adversarial attacks well, in particular under higher attack magnitudes and under none-gradient attacks.

\bibliographystyle{IEEEbib}
\bibliography{refs}

\end{document}